\documentclass[letterpaper, 10 pt, conference]{ieeeconf}  

\IEEEoverridecommandlockouts                              

\usepackage{amsmath,amssymb}
\usepackage{euscript}
\usepackage{subfig}
\usepackage{url}
\usepackage{hyperref}
\usepackage{mathrsfs}
\usepackage{array}
\usepackage{wrapfig}
\usepackage{graphicx} 

\usepackage{epsfig,float,alltt}
\usepackage{psfrag,xr}
\usepackage{pdfpages}
\usepackage{epstopdf}
\usepackage{pstool}
\usepackage{bm}
\usepackage[T1]{fontenc}
\usepackage{multirow}
\usepackage{balance}

\newcommand{\pos}{\bm{p}}

\usepackage{multirow}
\newcommand{\bom}    {\mbox{\boldmath $\omega$}}

{\end{tabbing} \end{mdframed} \vspace{5px}}

\newcommand{\BM}{\begin{bmatrix}}
\newcommand{\EM}{\end{bmatrix}}

\newcommand{\beq}{\begin{equation}}
\newcommand{\eeq}{\end{equation} }

\usepackage{booktabs}

\newcommand{\figref}[1]{Fig.~\ref{#1}}
\newcommand{\secref}[1]{Sec.~\ref{#1}}

\title{\LARGE \bf
Hierarchical Adaptive Loco-manipulation Control for Quadruped Robots }

\author{Mohsen Sombolestan and Quan Nguyen
\thanks{M. Sombolestan and Q. Nguyen are with the Department of Aerospace and Mechanical Engineering, University of Southern California, Los Angeles, CA 90089, email: {\tt somboles@usc.edu, quann@usc.edu}.}
}

\begin{document}

\maketitle
\thispagestyle{empty}
\pagestyle{empty}
\begin{abstract}

Legged robots have shown remarkable advantages in navigating uneven terrain. However, realizing effective locomotion and manipulation tasks on quadruped robots is still challenging. In addition, object and terrain parameters are generally unknown to the robot in these problems. Therefore, this paper proposes a hierarchical adaptive control framework that enables legged robots to perform loco-manipulation tasks without any given assumption on the object's mass, the friction coefficient, or the slope of the terrain. In our approach, we first present an adaptive manipulation control to regulate the contact force to manipulate an unknown object on unknown terrain. We then introduce a unified model predictive control (MPC) for loco-manipulation that takes into account the manipulation force in our robot dynamics. The proposed MPC framework thus can effectively regulate the interaction force between the robot and the object while keeping the robot balance. Experimental validation of our proposed approach is successfully conducted on a Unitree A1 robot, allowing it to manipulate an unknown time-varying load up to $7$ $kg$ ($60\%$ of the robot's weight). Moreover, our framework enables fast adaptation to unknown slopes or different surfaces with different friction coefficients. 

\end{abstract}

\section{Introduction} \label{sec:Introduction}

With a significant advantage in navigating rough terrain, legged robots can be suitable for applications in disaster rescue, the construction industry, last-mile delivery, or logistics. Such applications also require the capability of manipulating heavy packages.
There have also been successful mobile manipulation platforms using quadruped robots (e.g., ANYmal quadruped with an arm \cite{Chiu2022AManipulation,Sleiman2021AManipulation,Ferrolho2022RoLoMa:Arms} and Spot mini with an arm \cite{Zimmermann2021GoArm}). For a quadruped robot with a mounted robotic arm, an MPC approach is introduced in \cite{Sleiman2021AManipulation} to simultaneously control locomotion and manipulation (called \textit{loco-manipulation}). 
Instead of using a robotic arm, legged robots could also leverage the use of their body \cite{Rigo2023ContactControl} and legs \cite{Murooka2021HumanoidMaps, Wolfslag2020OptimisationRobots} to perform manipulation tasks. 
These approaches, however, are limited by the payload capacity of the robot arm.
In this paper, we are interested in leveraging the robot's body and locomotion to manipulate a heavy object.

A recent work \cite{Yang2022CollaborativeRobots} employs multiple quadruped robots for towing a load with cables to reach a target while avoiding obstacles.
In this work, as well as in manipulation in general, the controller often requires prior knowledge of the manipulated object and terrain, such as the object's mass and friction coefficient.
However, in many practical applications, the parameters of the manipulated object are generally unknown, and the robot should be able to adapt to a wide variety of objects.
Some previous work has applied adaptive control for collaborative manipulation in mobile robots without any assumption on the object's mass. They have developed centralized controllers \cite{Hu1995MotionTasks,Li2008RobustManipulators} as well as decentralized controllers \cite{Liu1998DecentralizedCooperations,Verginis2017RobustInformation,Culbertson2021DecentralizedBodies,Fink2008Multi-robotObstacles}. Nevertheless, in these works, it is assumed that the object is attached rigidly to the robots during the manipulation task. Moreover, measurement of manipulators' relative positions from the center of mass is sometimes required \cite{Prattichizzo2008Grasping}.


\begin{figure}[t!]
	\center
	\includegraphics[width=1\linewidth]{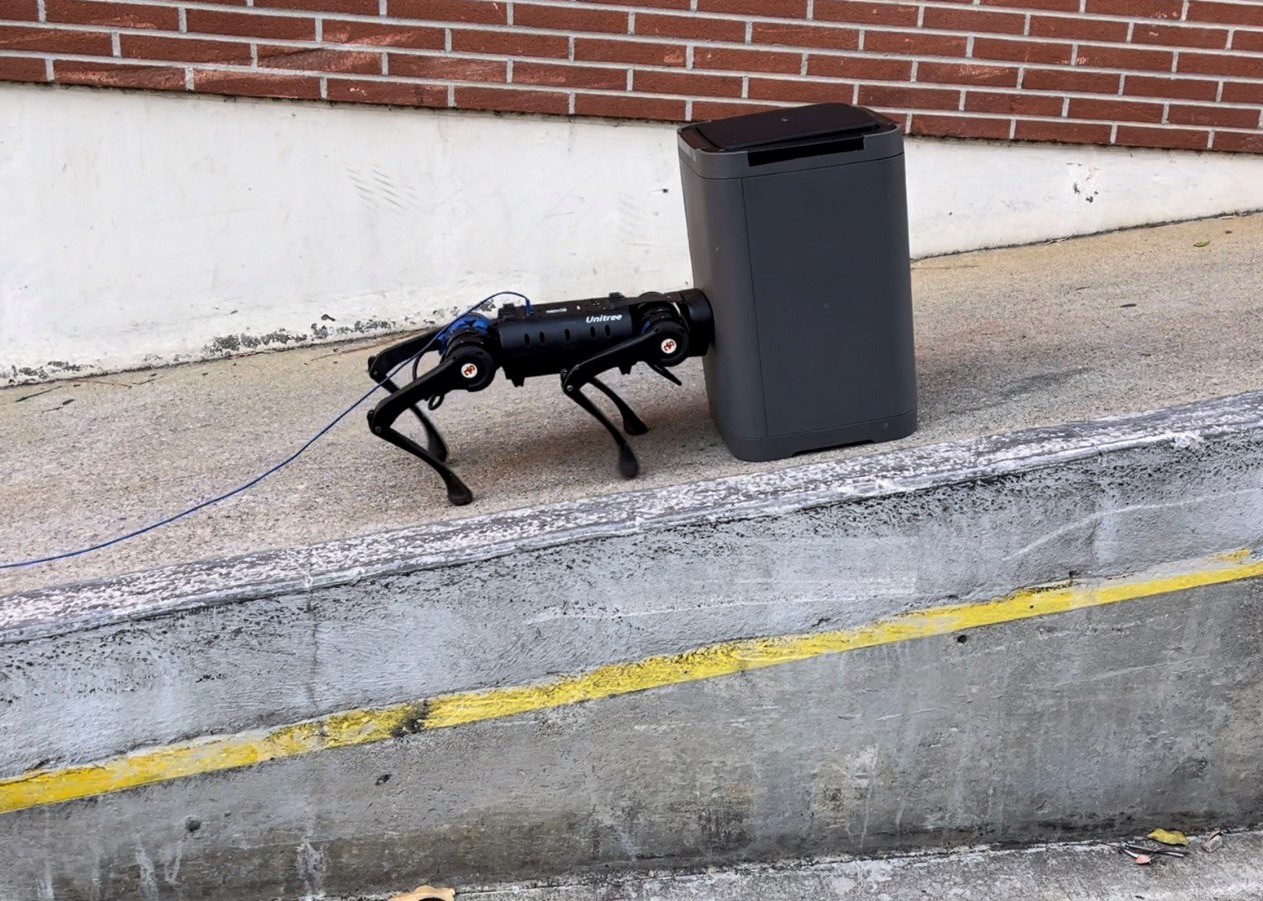}
	\caption{
Unitree A1 robot manipulating an unknown object of $5$ $kg$ on an unknown high-sloped terrain. Experimental results: \protect\url{https://youtu.be/-EvSmJRrMFI}.} 
	\label{fig:sloped_terrain}
	\vspace{-1em}
\end{figure}

The recent development of the model predictive control (MPC) for legged robots \cite{DiCarlo2018a, Li2021Force-and-moment-basedRobots} enables robust locomotion control with various gaits. Thanks to the capability of addressing dynamic constraints associated with friction, an MPC-based approach has also been implemented for robotic manipulators \cite{Hogan2020ReactiveControl}.
In our proposed approach, we develop a unified MPC framework to leverage robot locomotion to effectively manipulate a heavy object without losing robot balance. In addition, our method can adapt to a wide variety of unknown objects and terrain properties.
Our previous work \cite{Sombolestan2021} incorporated adaptive control into the force-based control framework to adapt to significant model uncertainty for legged locomotion. 
In this paper, we introduce a hierarchical adaptive control approach in combination with MPC to realize effective loco-manipulation under significant model uncertainty of object dynamics and terrain parameters. 
To the best of our knowledge, we are the first to approach solving the loco-manipulation task for quadruped robots without prior information about the manipulated object.

In our approach, we first introduce an adaptive control scheme to generate the interaction forces for the manipulation tasks. The controller will drive the object to follow the desired trajectory even under significant model uncertainty. 
Then, we will integrate the interaction force as the manipulation state into the MPC formulation we already had for locomotion control. Therefore, we will have a unified MPC framework for the loco-manipulation task that regulates the interaction forces for manipulation while maintaining robust locomotion.
Our approach is successfully validated in both high-fidelity simulations and hardware experiments. Although the conventional MPC for locomotion \cite{DiCarlo2018a} fails to manipulate objects toward the desired trajectory, our proposed hierarchical adaptive controller can effectively adapt to unknown time-varying loads and terrain parameters (e.g., slope and friction coefficient) with a minimal tracking error in the object motion. Thanks to this combination of controllers, our method can also allow the robot to climb an unknown slope while manipulating an object with an unknown mass of $5~kg$ (shown in \figref{fig:sloped_terrain}).

	
	
	
	
	

The remainder of the paper is organized as follows. \secref{sec:overview} presents an overview of the control system. The proposed method, including the design of the adaptive controller for manipulation and unified MPC, is elaborated in \secref{sec:Method}. Furthermore, the numerical and experimental validation are shown in \secref{sec:simulation} and \secref{sec:experiment}, respectively. Finally, \secref{sec:conclusion} provides concluding remarks.
\section{System Overview} \label{sec:overview}
\begin{figure}[t!]
	\center
	\includegraphics[width=1\linewidth]{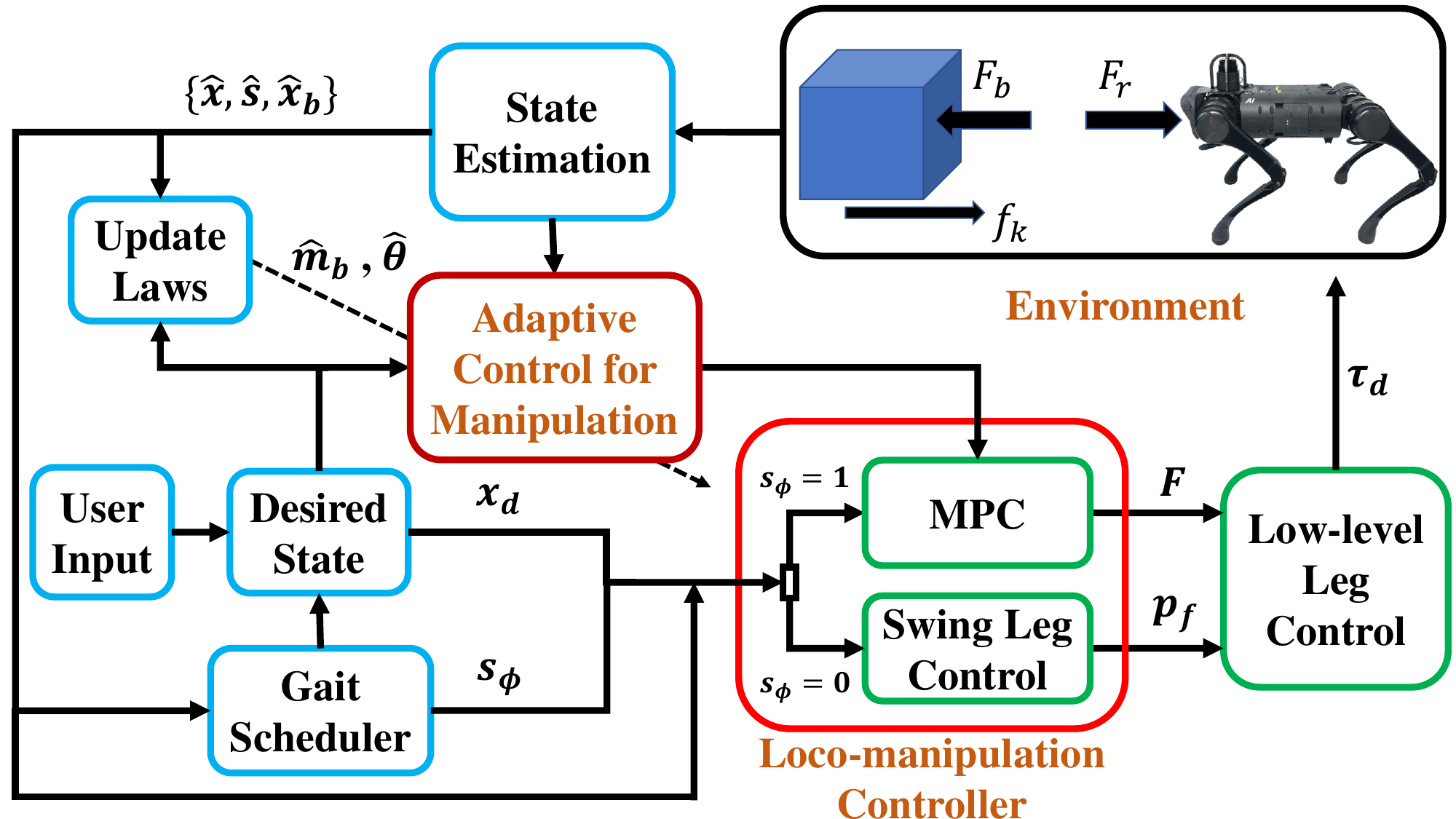}
	\caption{Block diagram of proposed hierarchical adaptive loco-manipulation control system}
	\label{fig:block_diagram}
	\vspace{-1em}
\end{figure}

Our proposed control architecture is illustrated in \figref{fig:block_diagram}. Our approach is based on a hierarchical adaptive control system to generate the required manipulation force. Then an MPC framework regulates the interaction force while keeping the robot balance. In this section, we will briefly introduce our approach and block diagram shown in \figref{fig:block_diagram}, then, in \secref{sec:Method}, we will elaborate on our proposed method.

The user defines appropriate input to generate the desired trajectory, including $xy$-velocity and yaw rate. Then, desired $xy$-position and yaw are determined by integrating the corresponding velocity. $z$ position contains a constant value of $0.3$ $m$, and the remaining desired states (roll, roll rate, pitch, pitch rate, and $z$-velocity) are always zero.

The gait scheduler utilizes independent boolean variables to define contact states scheduled $\bm{s}_{\phi} \in \{1 = \text{contact}, 0 = \text{swing}\}$ and switch each leg between swing and stance phases. Based on the contact schedule, the controller will execute either position control for swing legs ($\bm{p}_f$) or MPC for stance legs. More details on gait definition in the gait scheduler and swing leg controller can be found in \cite{Bledt2018,DiCarlo2018a}.

The state estimation includes contact estimation ($\hat{s}$), robot state ($\hat{x}$), and the manipulated object state ($\hat{x}_b$). However, obtaining the object state in a practical situation requires additional equipment for motion tracking. In \secref{sec:Method}, we will describe how to eliminate the object state estimation and instead use the robot state. 

First, a high-level adaptive controller generates the desired interaction force ($\bm{F}_b$) for the manipulation task. Since the object parameters such as mass ${m}_b$ and external force $\bm{f}_k$ are unknown, the adaptive controller exploits the estimated value  for mass $\hat{m}_b$ and vector of unknown parameters in external force $\hat{\bm{\theta}}$. The estimated parameters are governed by appropriate update laws to guarantee stable trajectory tracking. Then, the computed force from the adaptive controller will be integrated into the MPC to regulate the $\bm{F}_b$ while having robust locomotion. Now, the MPC solves the locomotion and manipulation control problem simultaneously. Finally, the ground reaction forces $\bm{F}$ achieved by MPC will be converted to the joint torques ($\bm{\tau}_d$) \cite{DiCarlo2018a}.

\section{Proposed Method - Hierarchical Adaptive  Loco-manipulation Control} \label{sec:Method}
The MPC introduced in \cite{DiCarlo2018a} considers only locomotion control. In this paper, we propose a hierarchical control system based on an adaptive controller to solve the locomotion and manipulation tasks simultaneously. First, we introduce an adaptive controller to generate force for object manipulation. Then we consider the computed force as a state in the MPC formulation and make a unified MPC for the loco-manipulation control. The unified MPC regulates the interaction force required for manipulation to achieve stable locomotion during the loco-manipulation task.

\subsection{Adaptive Control for Manipulation} \label{sec:Manipulation}
Let us consider the translational motion of a rigid object to be manipulated as shown in \figref{fig:block_diagram}. The linear motion is given by:
\begin{align}
\label{eq:newton}
\bm{F}_b = m_b \bm{\ddot{x}} + \bm{f}_k
\end{align}
where $\bm{F}_b \in \mathbb{R}^{3}$ is the applied force to the rigid object, $m_b$ and $\bm{\ddot{x}} \in \mathbb{R}^{3}$ are the mass and the acceleration of the rigid object, and $\bm{f}_k \in \mathbb{R}^{3}$ represents any external forces and nonconservative forces such as friction force. For example, when the object is on an $\alpha$-angle slope, the $\bm{f}_k$ also contains the projection of the object's weight along the slope surface ($m_b g \sin{\alpha}$).  
 
The mass of the rigid object ($m_b$) and the external force ($\bm{f}_k$) are unknown to the robot. The external forces can be rewritten as:
\begin{align}
\label{eq:friction force}
\bm{f}_k = \bm{Y}_f \bm{\theta}
\end{align}
where $\bm{Y}_f$ is the known regressor matrix and $\bm{\theta}$ is the vector of unknown parameters related to external forces.

A linear combination of position and velocity error $\bm{s}$ typically has been used in the adaptive controller for manipulators \cite{Slotine1991AppliedControl}, which can exhibit exponentially stable dynamics on the surface $\bm{s}=0$. Therefore, we define the following composite error:
\begin{align}
\label{eq:composite error}
\bm{s} = \dot{\bm{e}} + \lambda \bm{e}
\end{align}
which $\bm{e} = \bm{x} - \bm{x}_d$ and $\dot{\bm{e}} = \dot{\bm{x}} - \dot{\bm{x}}_d$ are linear tracking error and linear velocity error for the rigid object. 

Let us consider a Lyapunov function candidate as follows:
\begin{align}
\label{eq:lyapunov function}
V(t) = \frac{1}{2}(\bm{s}^T \bm{P} \bm{s} + {\tilde{m}_b}^{T} {\bm{\Gamma}_m}^{-1} \tilde{m}_b + \tilde{\bm{\theta}}^{T} {\bm{\Gamma}_f}^{-1} \tilde{\bm{\theta}})
\end{align}
where $\tilde{m}_b = \hat{m}_b - m_b$ is the mass estimation error with $\hat{m}_b$ being mass estimation. Similarly, $\tilde{\bm{\theta}} = \hat{\bm{\theta}} - \bm{\theta}$ is the external force parameter estimation error with $\hat{\bm{\theta}}$ being external force parameter estimation. 
$\bm{P}$, $\bm{\Gamma}_m$, and $\bm{\Gamma}_f$ are constant positive definite matrices, which correspond to adaptation gains. Using the fact that the estimation error derivatives $\dot{\tilde{m}}_b$, $\dot{\tilde{\bm{\theta}}}$ are the same as the estimation derivative $\dot{\hat{m}}_b$, $\dot{\hat{\bm{\theta}}}$, since the real values are constant, and by differentiating the $V(t)$, we will have:
\begin{align}
\label{eq:lyapunov function differential}
\dot{V}(t) = \bm{s}^T \bm{P} \dot{\bm{s}} + {\tilde{m}_b}^{T} {\bm{\Gamma}_m}^{-1} \dot{\hat{m}}_b + {\tilde{\bm{\theta}}}^{T} {\bm{\Gamma}_f}^{-1} \dot{\hat{\bm{\theta}}}.
\end{align}
We can expand the term 
\begin{align}
\label{eq:expand}
\bm{s}^T \bm{P} \dot{\bm{s}} = \bm{s}^T \bm{P} (\ddot{\bm{e}} + \lambda \dot{\bm{e}}) = \bm{s}^T \bm{P} [\ddot{\bm{x}} - ({\ddot{\bm{x}}}_d - \lambda \dot{\bm{e}})]
\end{align}
Let us define $\bm{Y}_m = {\ddot{\bm{x}}}_d - \lambda \dot{\bm{e}}$.
By assigning $\bm{P} = m_b \bm{I}_3$ and considering the equation \eqref{eq:newton}, the equation \eqref{eq:expand} can be rewritten as:
\begin{align}
\label{eq:expand final}
\bm{s}^T \bm{P} \dot{\bm{s}} = \bm{s}^T (\bm{F}_b - \bm{Y}_f \bm{\theta} - \bm{Y}_m m_b)
\end{align}
and Lyapunov function derivative would be:
\begin{align}
\label{eq:lyapunov function derivative}
\dot{V}(t) = & \bm{s}^T (\bm{F}_b - \bm{Y}_f \bm{\theta} - \bm{Y}_m m_b)  \\ \nonumber
&+ {\tilde{m}_b}^{T} {\bm{\Gamma}_m}^{-1} \dot{\hat{m}}_b + {\tilde{\bm{\theta}}}^{T} {\bm{\Gamma}_f}^{-1} \dot{\hat{\bm{\theta}}}.
\end{align}

Now, we will propose the control and adaptation laws. Let 
\begin{align}
\label{eq:control law}
\bm{F}_b = \bm{Y}_m \hat{m}_b + \bm{Y}_f \hat{\bm{\theta}} - \bm{K}_D \bm{s}
\end{align}
where $\bm{K}_D$ is a positive definite matrix. This control law contains terms related to estimated dynamics ($\bm{Y}_m \hat{m}_b + \bm{Y}_f \hat{\bm{\theta}}$) and a PD term ($\bm{K}_D \bm{s}$) which can lead the system to track the desired translational motion. Moreover, the proposed adaptation laws are:
\begin{align}
\label{eq:adaptation law 1}
\dot{\hat{m}}_b = -\bm{\Gamma}_m  {\bm{Y}_m}^T \bm{s} \\ \label{eq:adaptation law 2}
\dot{\hat{\bm{\theta}}} = -\bm{\Gamma}_f  {\bm{Y}_f}^T \bm{s}
\end{align}

\subsection{Stability Proof} \label{sec:stability}

Substituting control law \eqref{eq:control law} into Lyapunov function derivative \eqref{eq:lyapunov function derivative} yields
\begin{align}
\label{eq:proof 1}
\dot{V}(t) & = \bm{s}^T (\bm{Y}_f \tilde{\bm{\theta}} + \bm{Y}_m \tilde{m}_b) - \bm{s}^T \bm{K}_D \bm{s}  \nonumber \\
& + {\tilde{m}_b}^{T} {\bm{\Gamma}_m}^{-1} \dot{\hat{m}}_b + {\tilde{\bm{\theta}}}^{T} {\bm{\Gamma}_f}^{-1} \dot{\hat{\bm{\theta}}}.
\end{align}
and substituting adaptation laws \eqref{eq:adaptation law 1} and \eqref{eq:adaptation law 2} into \eqref{eq:proof 1}, yields
\begin{align}
\label{eq:proof 2}
\dot{V}(t) & =  - \bm{s}^T \bm{K}_D \bm{s} \leq 0
\end{align}
Since $\bm{V}(t)$ is positive definite and decrescent and $\dot{\bm{V}}(t)$ is negative semi-definite, the system is uniformly stable based on the Lyapunov theorem \cite{Slotine1991AppliedControl}. Therefore, $\bm{s}$, $\tilde{m}_b$, and $\tilde{\bm{\theta}}$ will be remained bounded.

From \eqref{eq:proof 2}, it can be perceived that $\bm{V}(t)$ has a finite limit. Moreover, it can be easily proven that $\dot{\bm{s}}$ is bounded \cite{Culbertson2021DecentralizedBodies}. Thus, from expression of $\ddot{\bm{V}}(t) = -2 \bm{s}^T \bm{K}_D \dot{\bm{s}}$, the $\ddot{\bm{V}}(t)$ is bounded. Now, since $\dot{\bm{V}}(t)$ is uniformly continuous in time ($\ddot{\bm{V}}(t)$ is bounded) and $\bm{V}(t)$ is lower bounded, then based on the second version of Barbalat’s Lemma \cite{Slotine1991AppliedControl}, $\dot{\bm{V}}(t) \rightarrow 0$ as $t \rightarrow \infty$. It implies that $\bm{s} \rightarrow 0$ as $t \rightarrow \infty$. When $\bm{s} = 0$ it can be obtained that $\dot{\bm{e}} = - \lambda \bm{e}$, which defines an asymptotically stable system.

\subsection{Unified MPC for Loco-manipulation Control}

In this subsection, we will introduce our proposed method for loco-manipulation control by integrating the manipulation force ($\bm{F}_b$) provided by adaptive control presented in \secref{sec:Manipulation} with the conventional MPC developed for quadrupeds' locomotion \cite{DiCarlo2018a}. The goal is to have a unified MPC formulation for quadrupeds to manipulate a rigid object with unknown parameters on unknown terrains while having robust locomotion.

$\bm{F}_b$ is the force that controls the rigid object manipulation, and $\bm{F}_r$ is the force exerted on the robot when manipulating a rigid object (see \figref{fig:block_diagram}). When the robot contacts the rigid object, we can assume that the robot and the rigid object are attached. Therefore, all the state measurements of rigid objects required for control law \eqref{eq:control law} (such as $\dot{\bm{e}}$, $\bm{e}$, and $\ddot{\bm{x}}_d$)  will be equally the same as the corresponding parameter of the robot. 
Moreover, it can be implied that the force applied to the rigid object is equal and opposite to the force applied to the robot, which means $\bm{F}_r = -\bm{F}_b$.
Thus, we can compute the $\bm{F}_r$ according to equations \eqref{eq:control law}, \eqref{eq:adaptation law 1}, and \eqref{eq:adaptation law 2}, by using robot states measurement. For consistency with \ref{sec:Manipulation}, the notation $-\bm{F}_b$ will be used instead of $\bm{F}_r$ throughout the rest of the paper.

Now, we can write the robot dynamic equation for the loco-manipulation control based on the state representation presented in \cite{DiCarlo2018a}:
\begin{align}\label{eq:loco-manipulation dynamic}
 \bm{\dot{X}} = \bm{D} \bm{X} + \bm{H} \bm{F} + \bm{F}_b / m
\end{align}
with
\begin{align}\label{eq:convenient SS components}
&\bm{X} = \left[\begin{array}{c} 
    \pos_{c} \\ 
    \bm{\Theta} \\ 
    \dot{\pos}_{c} \\ 
    \bom_{b}\\
    ||\bm{g}||
    \end{array} \right] \in \mathbb{R}^{13}
\\ \nonumber
&\bm{D} = \left[\begin{array}{ccccc} 
    \bm{0}_3 & \bm{0}_3 & \bm{I}_3 & \bm{0}_3 & \bm{0}_{3 \times 1}\\ 
    \bm{0}_3 & \bm{0}_3 & \bm{0}_3 & \bm{R}_z(\psi) & \bm{0}_{3 \times 1}\\ 
    \bm{0}_3 & \bm{0}_3 & \bm{0}_3 & \bm{0}_3 & \frac{\bm{g}}{||\bm{g}||}\\ 
    \bm{0}_3 & \bm{0}_3 & \bm{0}_3 & \bm{0}_3 & \bm{0}_{3 \times 1}\\
    \bm{0}_{1 \times 3} & \bm{0}_{1 \times 3} & \bm{0}_{1 \times 3} & \bm{0}_{1 \times 3} & 0
, \end{array} \right] \in \mathbb{R}^{13 \times 13} \\ \nonumber
&\bm{H} = \left[\begin{array}{ccc} 
\bm{0}_{3} & \dots & \bm{0}_{3} \\
\bm{0}_{3} & \dots & \bm{0}_{3} \\
  \mathbf{I}_{3} / m & \dots & \mathbf{I}_{3}/m \\
    \hspace{-0.1cm}  \bm{I}_G^{-1} [\bm{p}_{1} - \pos_{c}] \times & \dots & \bm{I}_G^{-1} [\bm{p}_{4} - \pos_{c}] \times \\
    \bm{0}_{1 \times 3} & \dots & \bm{0}_{1 \times 3}
  \end{array} \right] \in \mathbb{R}^{13 \times 12}
\end{align}
where $m$ is the robot's mass, $\bm{I}_G\in \mathbb{R}^{3 \times 3}$ is the moment of inertia in the world frame, $\bm{g}\in \mathbb{R}^{3}$ is the gravity vector, $\bm{R}_z(\psi)$ is the rotation matrix corresponding to the yaw angle $\psi$, $\bm{p}_{c}\in \mathbb{R}^{3}$ is the position of the center of mass (COM), and $\bm{p}_{i}\in \mathbb{R}^{3}$ ($i \in \{1,2,3,4\}$) are the positions of the feet, $\ddot{\pos}_{c}\in \mathbb{R}^{3}$ is body’s linear acceleration, $\dot \bom_b\in \mathbb{R}^{3}$ is angular acceleration, and $\bm{F} = [\bm{F}_1^T, \bm{F}_2^T, \bm{F}_3^T, \bm{F}_4^T]^T \in \mathbb{R}^{12}$ are the ground reaction forces acting on each of the robot’s four feet. Similar to what has been shown in \figref{fig:block_diagram}, the $\bm{F}_r$ is applied to the robot's head, which is approximately along the body's center of mass. Hence, we can neglect the moment resulting from $\bm{F}_r$ around the robot's center of mass in equation \eqref{eq:loco-manipulation dynamic}.

Since linear MPC will predict the dynamic over a finite time horizon, it requires linear discrete-time dynamics. 
However, to employ a conventional discretization method such as zero-order hold, the manipulation term $\bm{F}_b/m$ in \eqref{eq:loco-manipulation dynamic} must be combined into the state vector in order to create an augmented vector for MPC formulation. Hence, the equation \eqref{eq:loco-manipulation dynamic}, can be written as follows:
\begin{align}
\label{eq:combined SS}
\bm{\dot{\eta}} = \bar{\bm{D}} \bm{\eta} + \bar{\bm{H}} \bm{F} 
\end{align}
where 
\begin{align} \label{eq:extended SS components}
&\bm{\eta} = \left[\begin{array}{c} 
    \bm{X} \\ 
    \hline
    \bm{F}_b /m
    \end{array} \right] \in \mathbb{R}^{16}
\\ \nonumber
&\bar{\bm{D}} = \left[\begin{array}{@{}c|c@{}} 
        \begin{matrix}
           \bm{D}_{13 \times 13}
        \end{matrix}
        & \begin{matrix}
            \bm{0}_{6 \times 3} \\
            \bm{I}_{3 \times 3} \\
            \bm{0}_{4 \times 3} \\
        \end{matrix}
    \\
    \hline
    \bm{0}_{3 \times 13} & \bm{0}_{3 \times 3}
\end{array} \right] \in \mathbb{R}^{16 \times 16}
\\ \nonumber
&\bar{\bm{H}} = \left[\begin{array}{c}
    \bm{H} \\
    \hline
    \bm{0}_{3 \times 12}
\end{array} \right] \in \mathbb{R}^{16 \times 12}
\end{align}
where $\bm{\eta}$ is the augmented vector. Therefore, a linear MPC can be designed as follows:
\begin{align}
\label{eq:mpc}
\min_{\bm{F}} \quad & \sum_{i=0}^{k-1} {\tilde{\bm{X}}_{i+1}}^T \bm{Q} {\tilde{\bm{X}}_{i+1}} + {\bm{F}_{i}}^T \bm{R} \bm{F}_{i} \\ \nonumber
\textrm{s.t.} \quad & \tilde{\bm{X}}_{i+1} = \bm{X}_{i+1} - \bm{X}_{i+1,d} \\ \nonumber
& \bm{\eta}_{i+1} = \bar{\bm{D}}_{t,i} \bm{\eta}_{i} + \bar{\bm{H}}_{t,i} \bm{F}_{i} \\ \nonumber
& \underline{\bm{d}} \leq \bm{C} \bm{F}_{i} \leq \bar{\bm{d}}
\end{align}
where $k$ is the number of horizons, $\bm{X}_{i,d}$ is the system desired state at time step $i$, $\bm{F}_{i}$ is the computed ground reaction forces at time step $i$, $\bm{Q}$ and $\bm{R}$ are diagonal positive semi-definite matrices, $\bar{\bm{D}}_{t,i}$ and $\bar{\bm{H}}_{t,i}$ are discrete-time system dynamics matrices.
The $\underline{\bm{d}} \leq \bm{C} \bm{F}_i \leq \bar{\bm{d}}$ represents friction cone constraints which are defined in \cite{Focchi2017a}.

\section{Numerical Simulation} \label{sec:simulation}
In this section, we validate the effectiveness of the proposed method in a high-fidelity simulation of the A1 robot from Unitree. The controller is implemented in ROS, and we use Gazebo as the simulator. In simulations, we want to  show an A1 robot manipulates an unknown object while trying to adapt to terrain uncertainty. To this end, we construct multiple simulation scenarios with two perspectives: 1) Adapt to object uncertainty, 2) Adapt to terrain uncertainty.

In these simulations, the robot manipulates an object in one direction (e.g., along the $x$-axis). So, since we simulate our model for one direction (1-D) manipulation, we can ignore the two other components ($yz$-direction) of $\bm{F}_b$. Therefore, for adaptive controller, the design parameters become scalar, and we set the parameters as follow: $\lambda = 2$, $\bm{K}_D = 200$, $\Gamma_m = 10$, and $\Gamma_f = 10$. In addition, all the estimated parameters ($\hat{m}_b$, $\hat{\bm{\theta}}$) start from zero. 
Note that the manipulation force $\bm{F}_b$ is a 3-D vector in general. Although we consider the robot manipulating an object along one axis in our implementation, the framework developed in \secref{sec:Method} is not limited to 1-D manipulation problems.

\subsection{Adapt to Object Uncertainty}
\begin{figure}[tb!]
	\centering
	\subfloat[Simulation snapshot of the A1 robot while manipulating an unknown $5$ $kg$ object]{\includegraphics[width=0.8\linewidth]{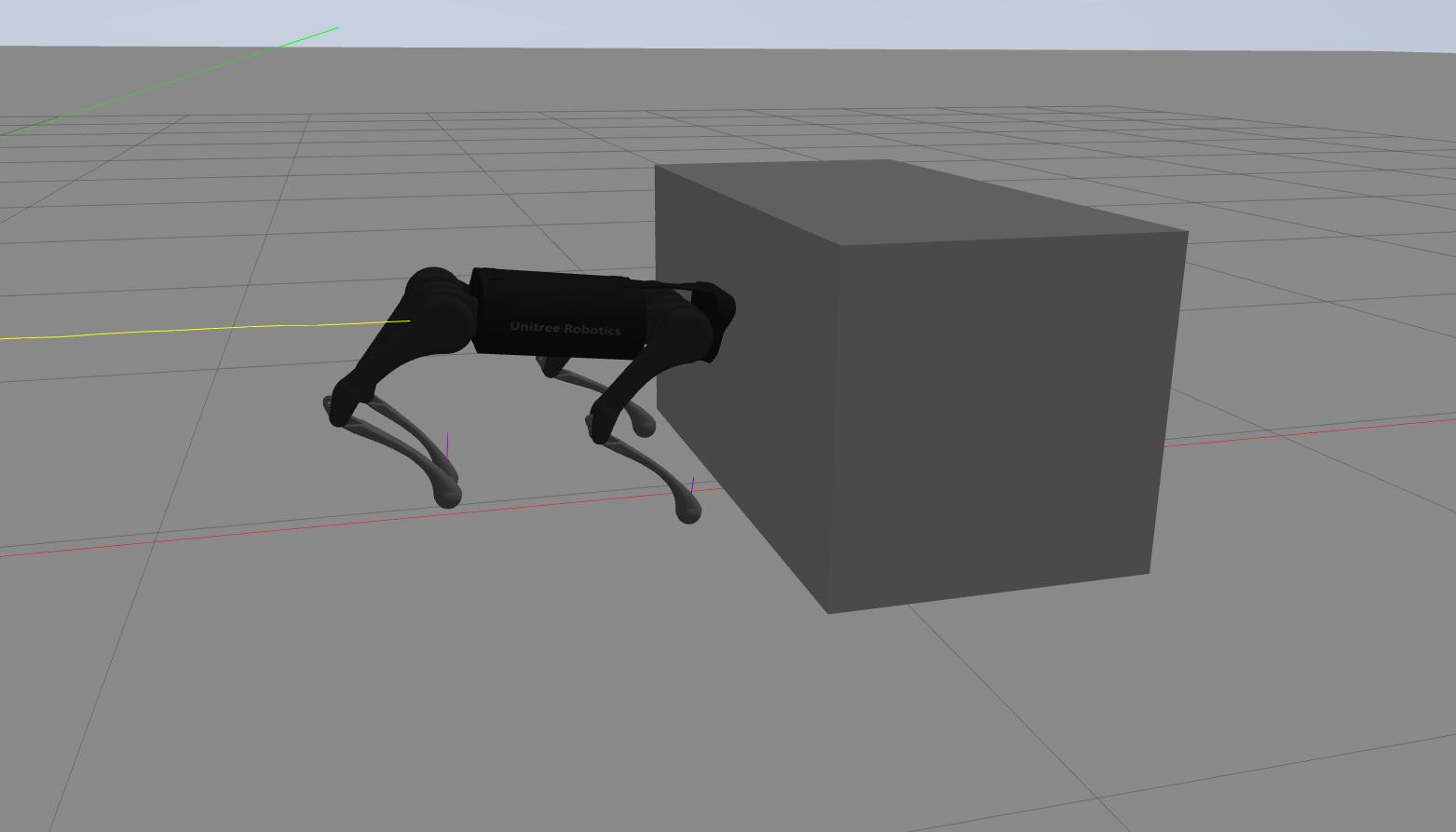}}
	\hfill
	\subfloat[Velocity tracking for a $3$  $kg$ object]{\includegraphics[width=1\linewidth]{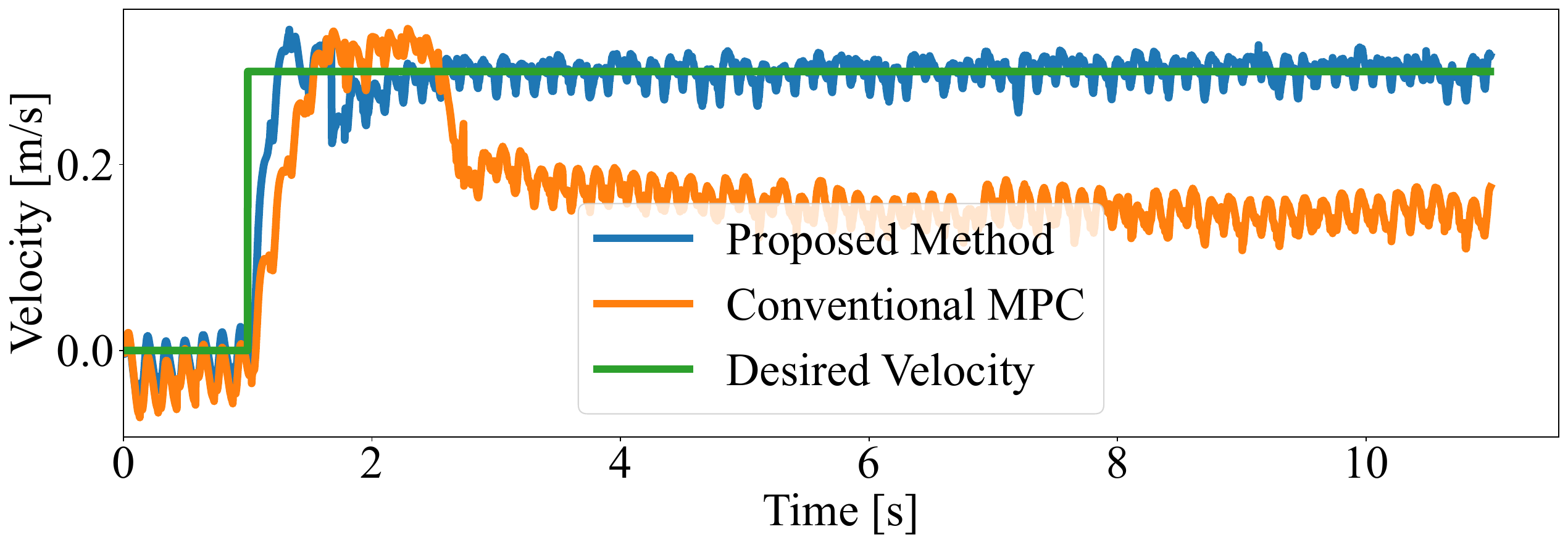}}
    \hfill
    \subfloat[Velocity tracking for a $5$ $kg$ object]{\includegraphics[width=1\linewidth]{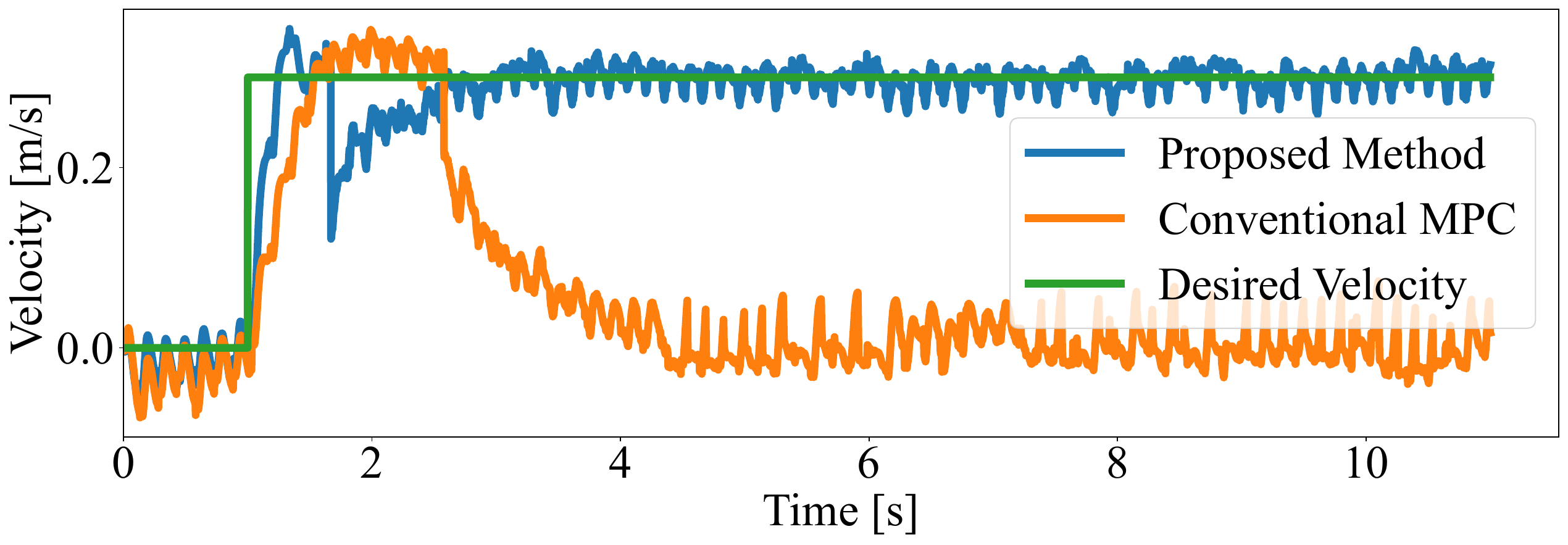}\label{subfig: velocity_tracking}}
	\hfill
	\subfloat[Mass estimation for $5$ $kg$ object]{\includegraphics[width=1\linewidth]{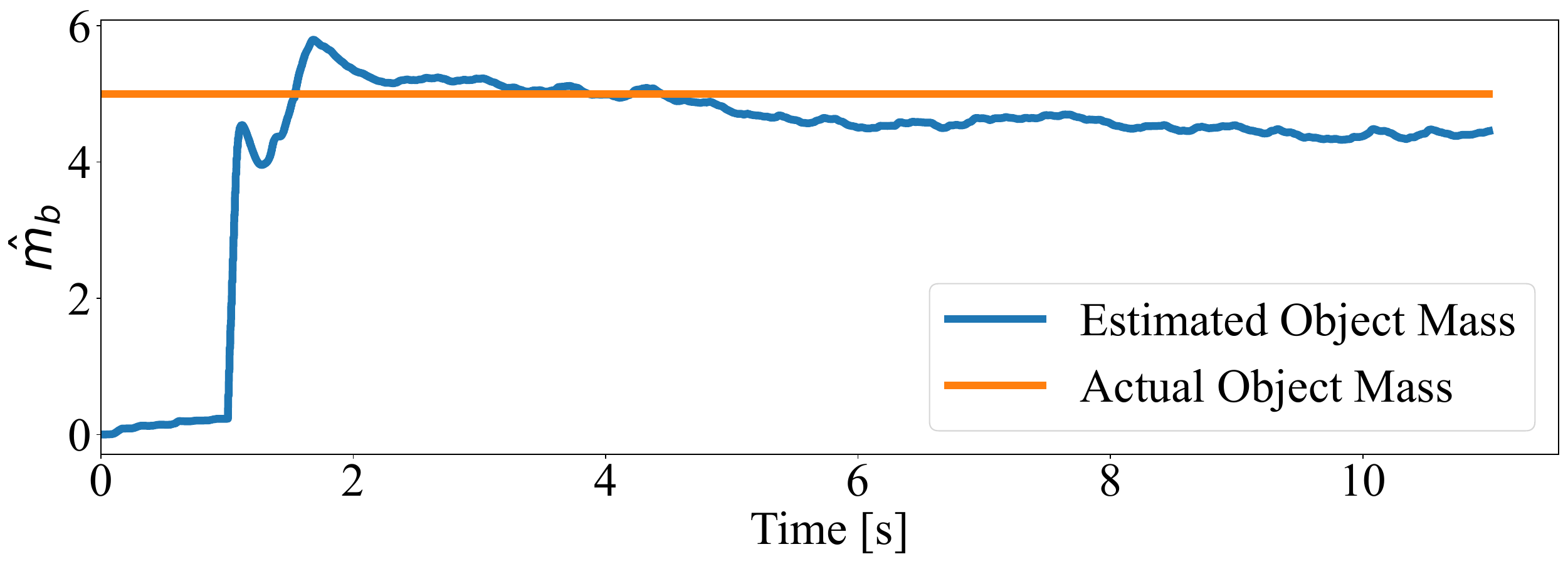}\label{subfig:estimated_mass}}
	
	\caption{Comparing simulation results of the proposed method and conventional MPC for the robot while manipulating an unknown object.}
	\label{fig:comparing_result}
		\vspace{-1em}
\end{figure}

First, we will compare the performance of our proposed method with the conventional MPC presented in \cite{DiCarlo2018a} to verify the effectiveness of our framework. We tested our controller when the robot tries to push an object of $3$ $kg$ and $5$ $kg$ with the desired velocity of $0.3$ $m/s$. The friction coefficient between the ground and the object is $0.6$. All the parameters related to the object's inertia and geometry are unknown to the robot. As shown in \figref{fig:comparing_result}, while the conventional MPC fails to track the desired velocity, our proposed method shows an accurate tracking result.

The mass estimation for the $5$ $kg$ object is also provided in \figref{subfig:estimated_mass}. As shown in \figref{fig:comparing_result}, the mass estimation error will be maintained within the small range. Remember equation \eqref{eq:proof 2} in \secref{sec:stability}, we proved that the mass estimation error ($\tilde{m}_b$) will remain bounded. In addition, the magnitude of the error can be reduced by increasing the adaptation gain $\bm{\Gamma}_m$. By increasing the $\bm{\Gamma}_m$, the control signals will be updated faster; however, for the reliability and robustness of the control scheme, it is essential to obtain smooth control signals. Comparing \figref{subfig: velocity_tracking} and \figref{subfig:estimated_mass} indicates that the estimated mass increases when the robot starts walking, even before contacting the object. 
This is plausible due to the error in velocity tracking at the start of walking since the adaptation law \eqref{eq:adaptation law 1} depends on the composite error $\bm{s}$. 

\subsection{Adapt to Terrain Uncertainty}
\begin{figure}[tb!]
	\centering
	\subfloat[Hardwood ground]{\includegraphics[width=0.48\linewidth]{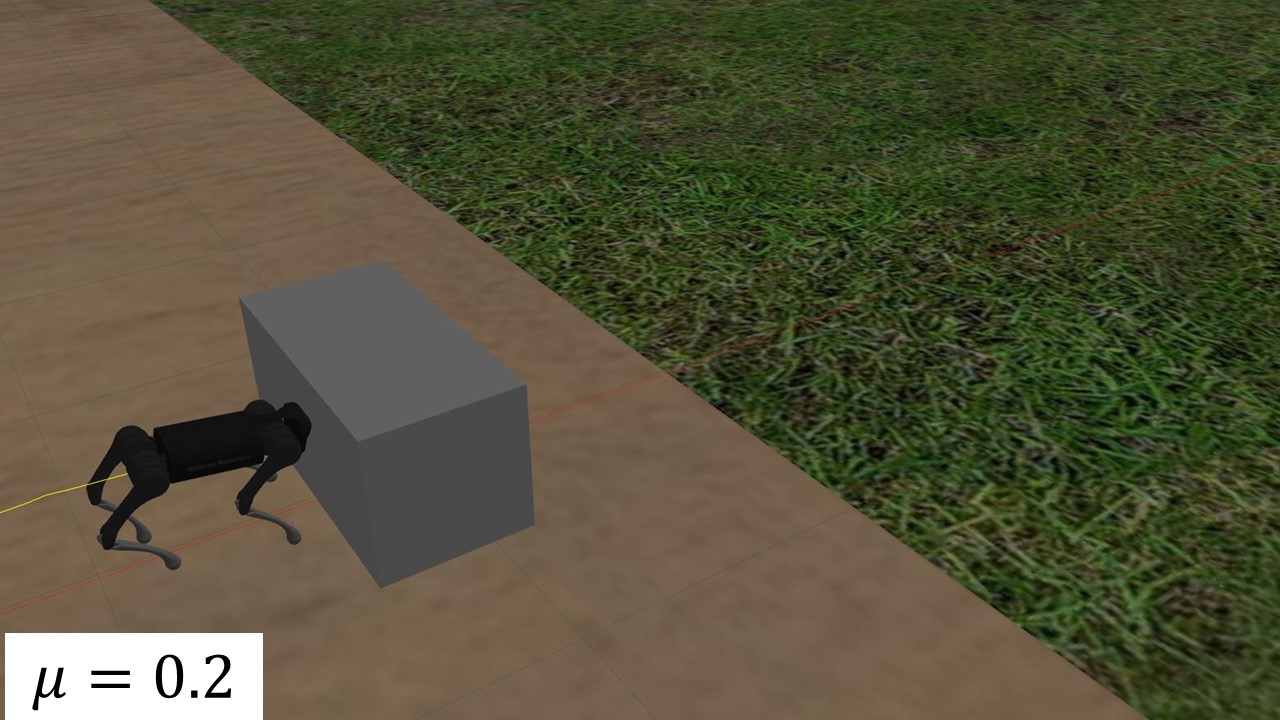}}
	\hfill
	\subfloat[Grass field]{\includegraphics[width=0.48\linewidth]{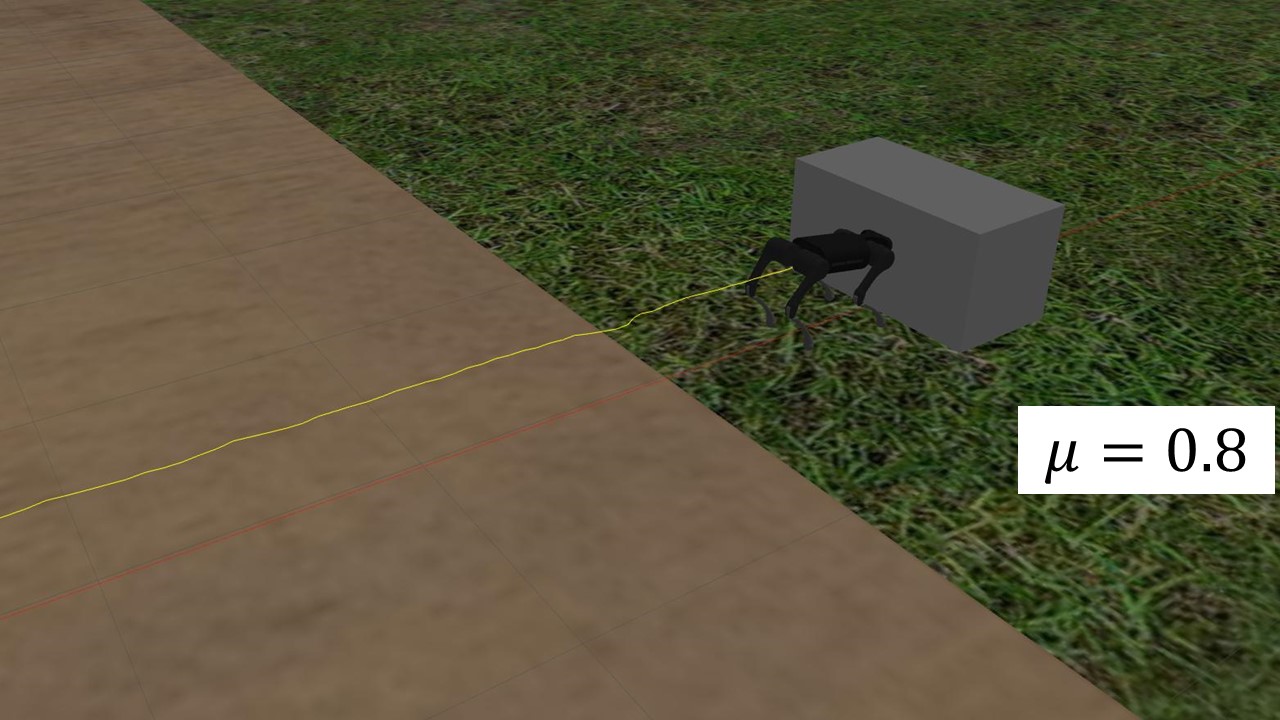}}
    \hfill
    \subfloat[Velocity tracking for an unknown $5$ $kg$ object]{\includegraphics[width=1\linewidth]{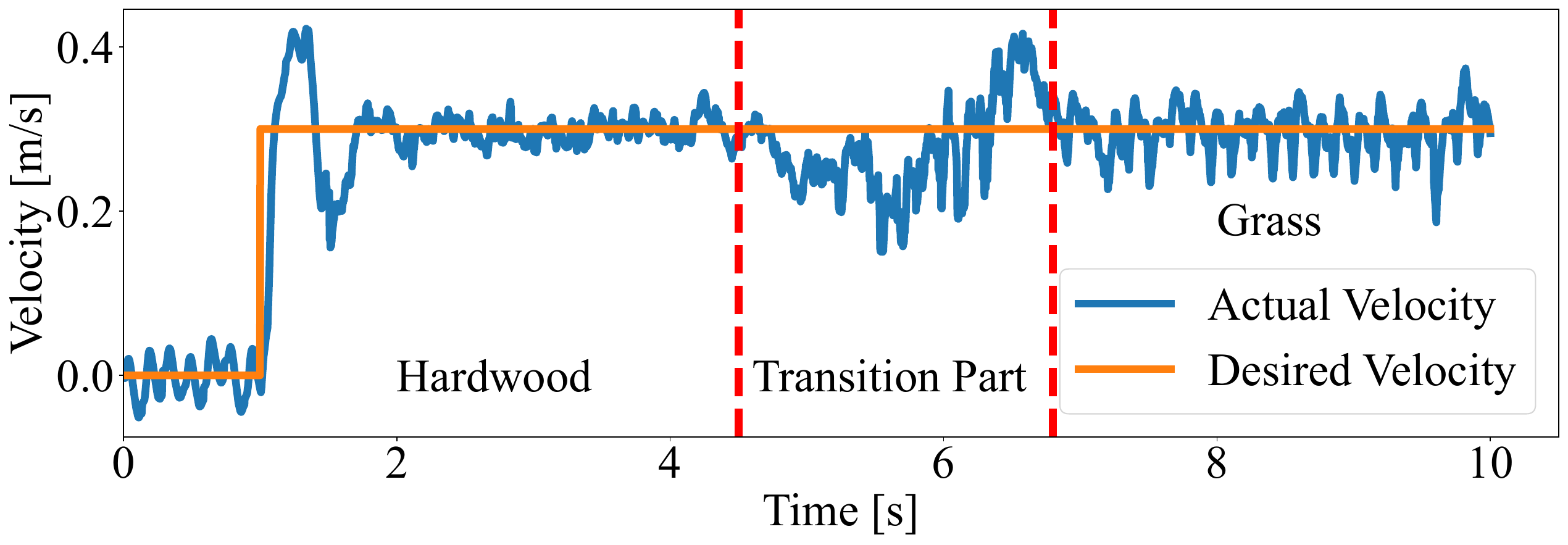}\label{subfig:velocity_tracking}}

	\caption{{\bfseries{Navigating surfaces with different friction properties.}} The transition part indicates when the robot tries to cross the hardwood ground to the grass field.}
	\label{fig:various_friction}
		\vspace{-1em}
\end{figure}

We aim to examine the robot's capability to adapt to terrain uncertainty. To this end, we simulate the robot navigating various terrain with different friction properties while pushing an unknown $5$ $kg$ object with the desired velocity of $0.3$ $m/s$. Some simulation snapshots are presented in \figref{fig:various_friction}. The robot starts from a hardwood ground with a friction coefficient of $0.3$; then, it passes across a grass field with a friction coefficient of $0.8$. The robot's velocity is depicted in \figref{subfig:velocity_tracking}. According to \figref{subfig:velocity_tracking}, when the robot tries to transit from hardwood ground to grass, the tracking error increases until the robot entirely moves on the grass. In the transition part, the object is on the grass, so a greater force is required for manipulation. However, the robot's feet are still on the hardwood, and due to the small friction on the hardwood, the robot cannot exert enough force for object manipulation.
\section{Hardware Experiment} 
\label{sec:experiment}
We successfully implemented our proposed method on robot hardware. In experiments, similar to simulations, we try to follow two perspectives: 1) Adapt to object uncertainty, and 2) Adapt to terrain uncertainty. To examine these viewpoints, we tested the A1 robot by manipulating a time-varying load and climbing a high-sloped terrain while pushing an object. More details of the robot's object manipulation are shown in the supplemental video\footnote{\url{https://youtu.be/-EvSmJRrMFI}}.

\subsection{Adapt to Object Uncertainty}

\begin{figure}[tb!]
	\centering
	\subfloat[Loading]{\includegraphics[width=0.48\linewidth]{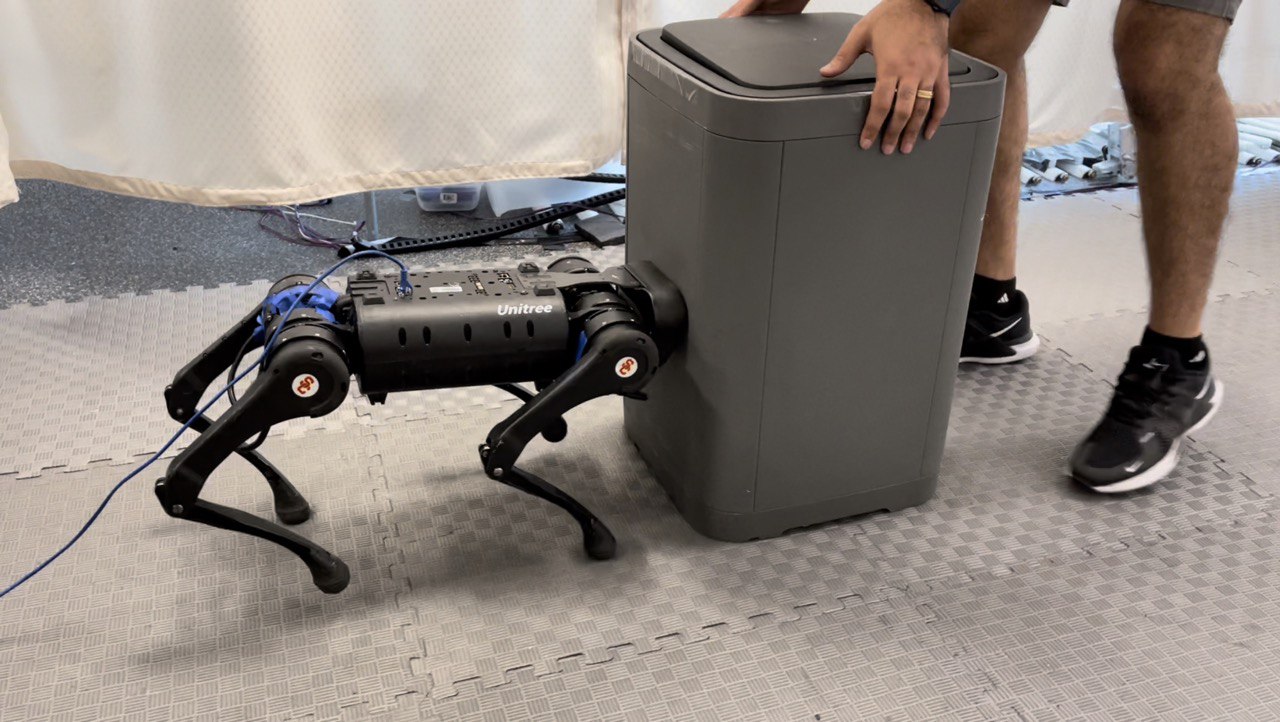}}
	\hfill
	\subfloat[Unloading]{\includegraphics[width=0.48\linewidth]{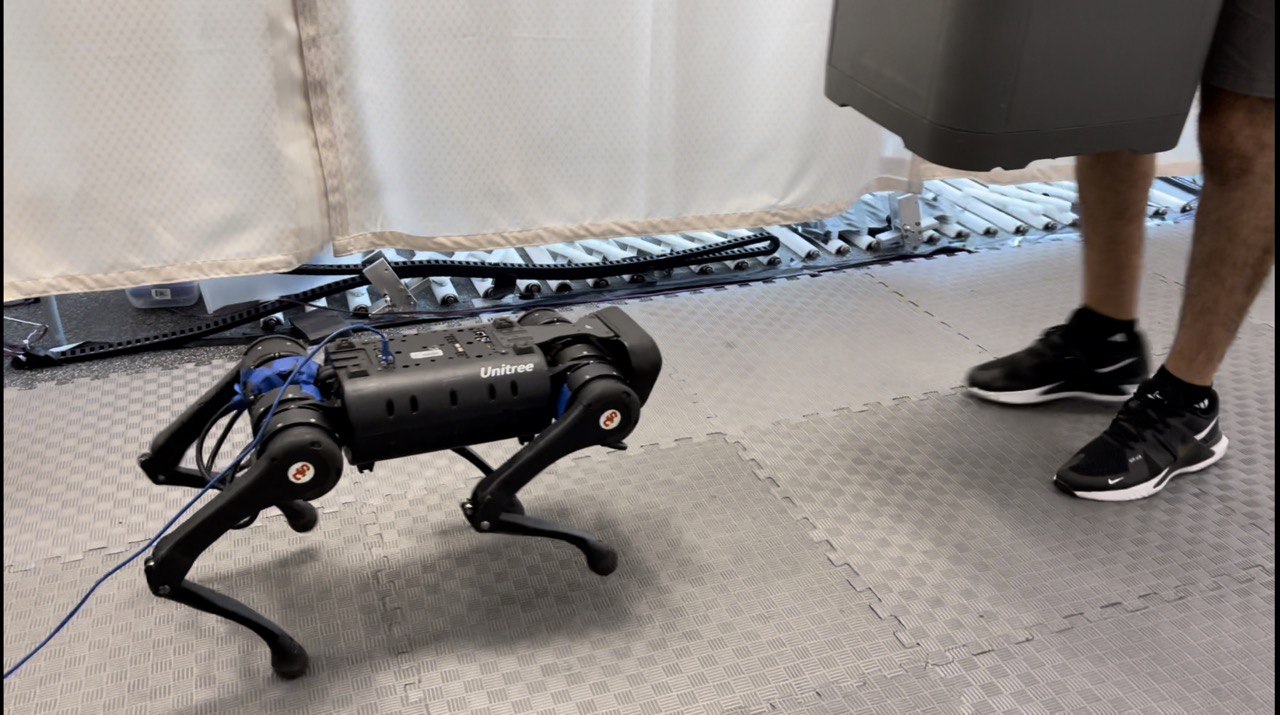}}
    \hfill
    \subfloat[Velocity tracking for an unknown $5$ $kg$ object]{\includegraphics[width=1\linewidth]{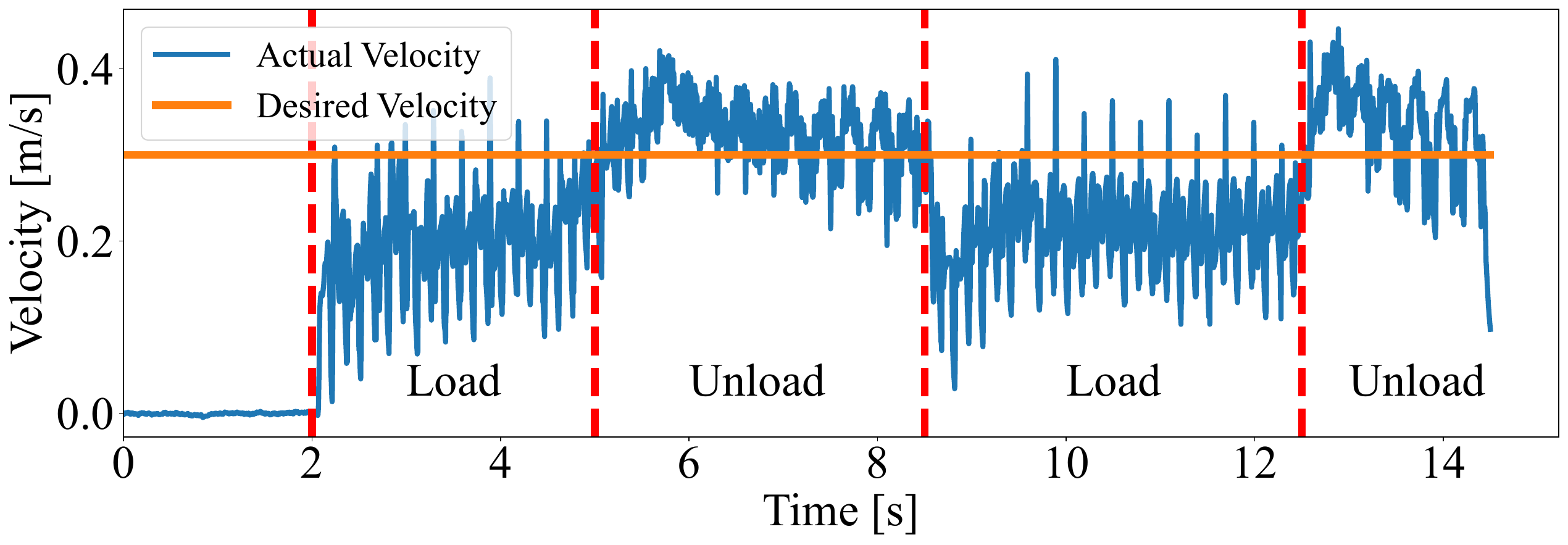}}

	\caption{Results for load/unload experiment.}
	\label{fig:load_unload}

\end{figure}

To verify the effectiveness of our method, we designed two experiments with varying object mass. For the first experiment, the robot starts by confronting an object for manipulation. After a while, we remove the object, and the robot continues walking normally. Again, after a few seconds, we put the object in front of the robot and made the robot manipulate the box again. The successive load/unload experiment demonstrates that the method is not necessarily restricted to loco-manipulation control and can handle locomotion only as well. Moreover, it is not an obligation to have a rigid connection between the robot and the object during the whole loco-manipulation operation. The velocity tracking for the experiment is presented in \figref{fig:load_unload}.

\begin{figure}[tb!]
	\centering
	\subfloat[Start point]{\includegraphics[width=0.48\linewidth]{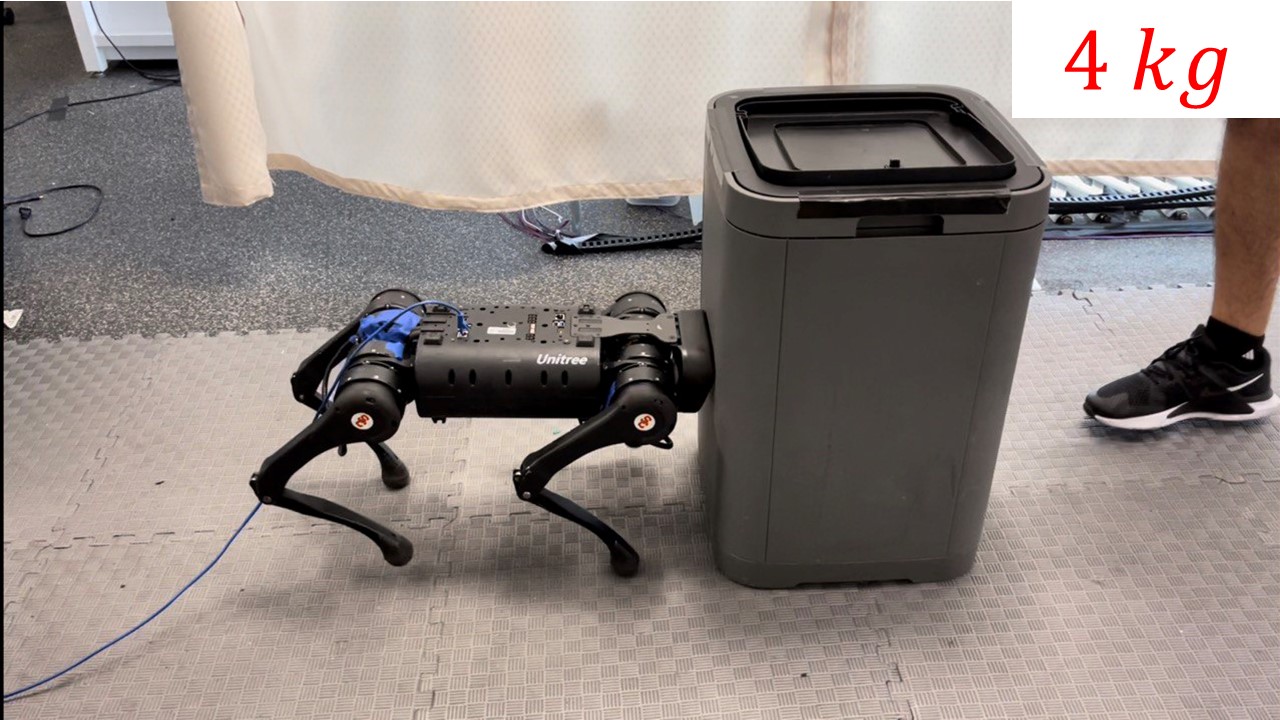}}
	\hfill
	\subfloat[Final point]{\includegraphics[width=0.48\linewidth]{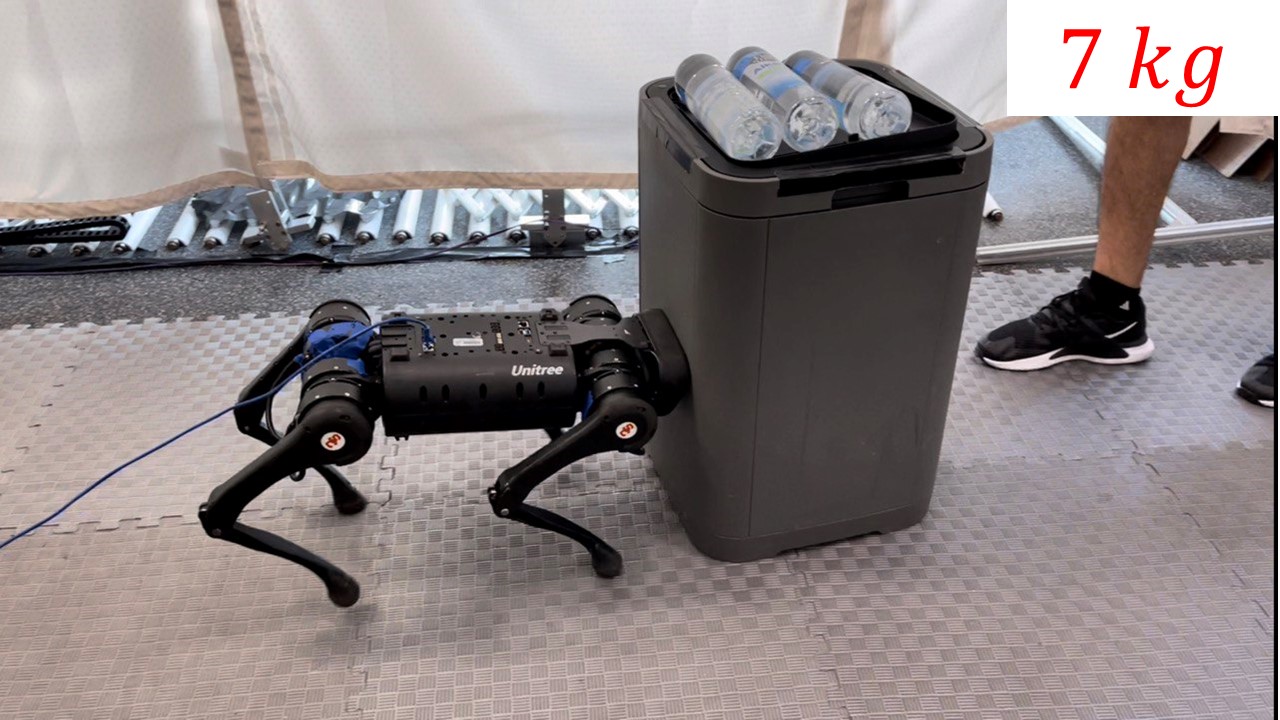}}
    \hfill
    \subfloat[Velocity tracking for an unknown time-varying load (from $4$ $kg$ to $7$ $kg$)]{\includegraphics[width=1\linewidth]{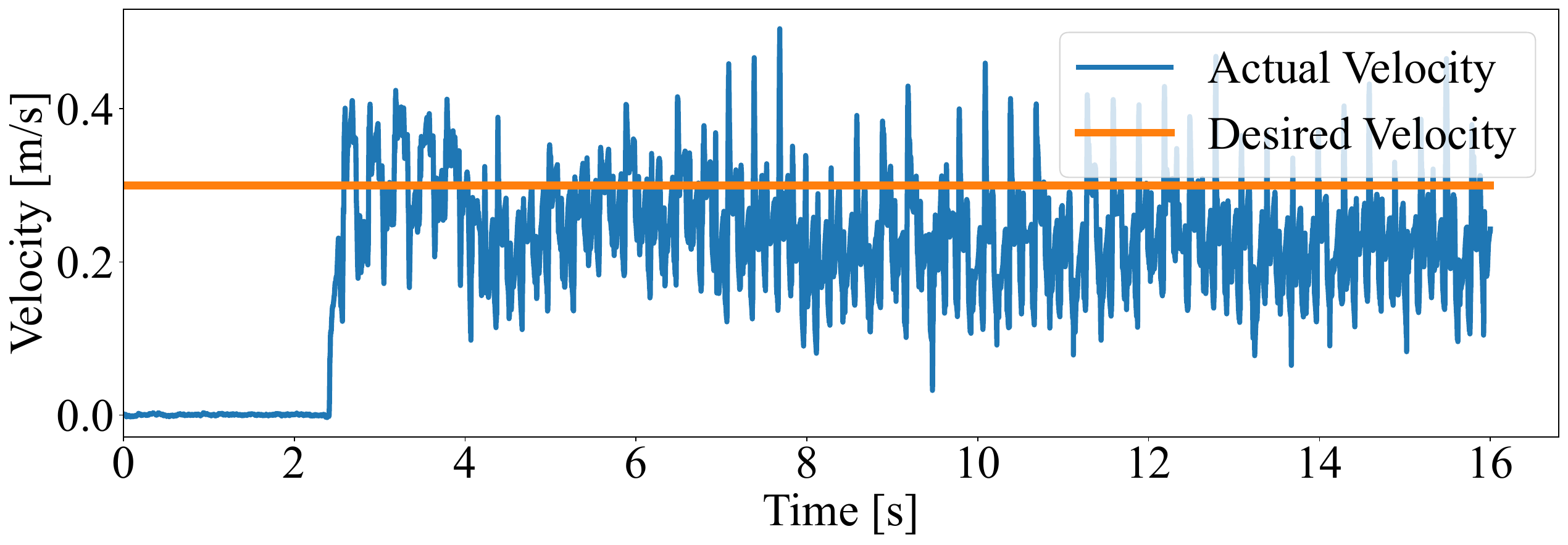}\label{subfig:velocity_time_varying}}

	\caption{Results for manipulating an unknown time-varying load up to $7$ $kg$.}
	\label{fig:time_varying_exp}
		\vspace{-1em}
\end{figure}

Furthermore, we tested the robot with a time-varying mass object. The robot starts to push a $4$ $kg$ object, and then we will add three more $1$ $kg$ water bottles sequentially. The velocity result is presented in \figref{fig:time_varying_exp}. According to \figref{subfig:velocity_time_varying}, although the object mass changes during the experiment, the robot has a smooth velocity, and the plot shows that the robot can adapt to the object uncertainty online.

\subsection{Adapt to Terrain Uncertainty}

\begin{figure}[tb!]
	\centering
	\subfloat[Locomotion only]{\includegraphics[width=0.48\linewidth]{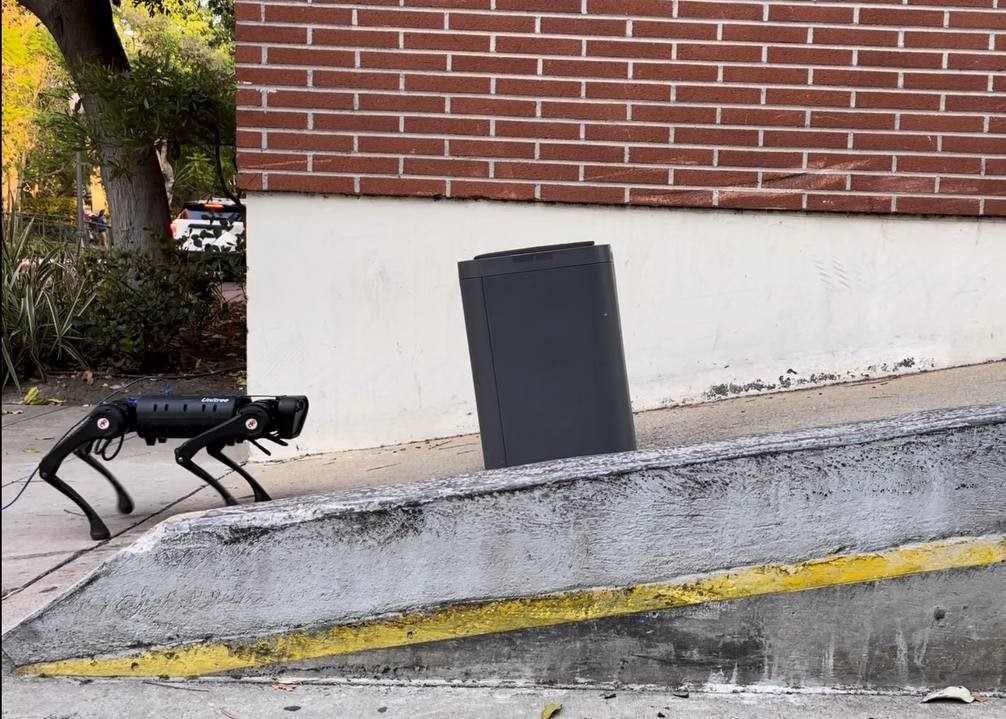}}
	\hfill
	\subfloat[Loco-manipulation]{\includegraphics[width=0.48\linewidth]{Figures/sloped_terrain_experiment/first_fig_updated.jpg}}
    \hfill
    \subfloat[The velocity of the robot along the slope]{\includegraphics[width=1\linewidth]{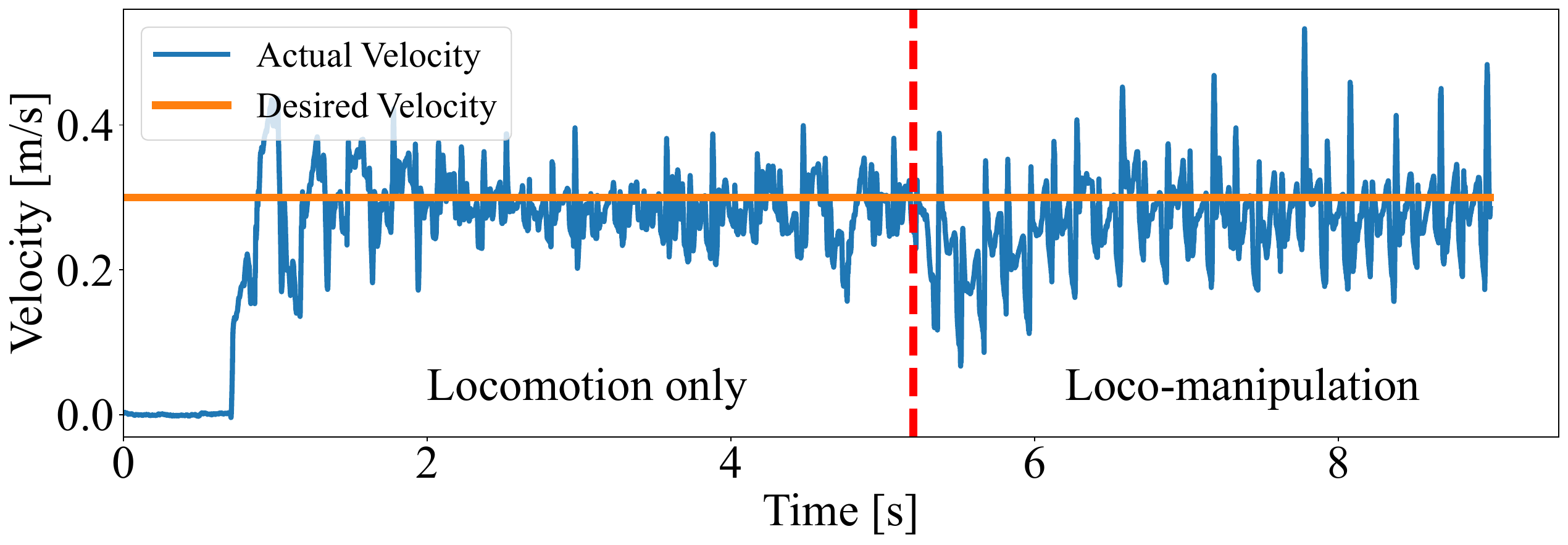}\label{subfig:slope_velocity}}

	\caption{Results for manipulating an unknown $5$ $kg$ object on a high-sloped terrain.}
	\label{fig:sloped_terrain_exp}
		\vspace{-1em}
\end{figure}

To examine the robot's adaptation to terrain uncertainty, we tested the robot to manipulate an object on a high-sloped terrain. The robot tries to push an unknown $5$ $kg$ object on the slope. The results are presented in \figref{fig:sloped_terrain_exp}. The plot in \figref{subfig:slope_velocity} shows the robot's velocity along the slope. The experiment starts with robot locomotion only, then the robot reaches the object and manipulates it along the slope.

While the object is on the slope, an additional term ($m_b g \sin{\alpha}$) will be added to the dynamic equation, representing the projection of the object's weight along the slope surface. This term can be considered as an external force ($\bm{f}_k$) in equation \eqref{eq:newton}. Thus, the adaptive controller designed in \secref{sec:Manipulation} can handle the terrain uncertainty without having any information about the slope angle, and the robot is capable of manipulating the object while climbing the slope. However, to adjust the robot's pitch angle on the slope, we estimate the slope angle based on the foot placement measurements. By considering the robot's front and rear feet position along the $x$-axis and $z$-axis, we can adjust the pitch angle to make the robot's body parallel to the slope. Again, note that the controller does not have any information about the slope's angle and friction properties between the object and the slope.


\section{Conclusion} \label{sec:conclusion}
In summary, we have presented a hierarchical adaptive control approach for quadruped robots to manipulate unknown objects while maintaining robot balance. We designed an adaptive controller to generate appropriate force commands for the manipulation task; then, we introduced a unified MPC that simultaneously considers locomotion and manipulation control. We have demonstrated the effectiveness of our method using numerical and experimental validations. The robot can manipulate an unknown time-varying load up to $7$ $kg$. Additionally, it can push an unknown $5$ $kg$ object and climb a slope while maintaining accurate trajectory tracking.
In contrast, the baseline MPC fails even to move the manipulated object. Moreover, our approach has shown that the robot can navigate the terrain with multiple friction coefficients. Therefore, our proposed method not only can compensate for the object uncertainty but also can adapt to unknown terrain properties.

In the future, we aim to extend our framework to two- and three-dimensional manipulation tasks. We also plan to develop hierarchical adaptive control for the collaborative manipulation of rigid objects using multiple quadrupeds.

\section*{Acknowledgments}
This work is supported in part by National Science Foundation Grant IIS-2133091. The opinions expressed are those of the authors and do not necessarily reflect the opinions of the sponsors.

\balance

\bibliographystyle{IEEEtranS}
\bibliography{references}

\end{document}